\documentclass{article}

\usepackage[numbers]{natbib}

 \usepackage[preprint]{neurips_2025}

\usepackage[utf8]{inputenc} 
\usepackage[T1]{fontenc}    
\usepackage{hyperref}       
\usepackage{url}            
\usepackage{booktabs}       
\usepackage{amsfonts}       
\usepackage{nicefrac}       
\usepackage{microtype}      
\usepackage{xcolor} 
\usepackage{graphicx} 
\usepackage{amsmath}
\usepackage{cleveref}
\usepackage{multirow} 
\usepackage{tabularx}
\usepackage{colortbl}
\usepackage{fontawesome}
\usepackage{placeins}
\definecolor{bestgreen}{RGB}{220,245,220}
\definecolor{worstred}{RGB}{255,225,225}
\newcommand{\best}[1]{\cellcolor{worstred}\textbf{#1}}
\newcommand{\worst}[1]{\cellcolor{bestgreen}#1}

\title{CLIP-Guided Label-Free Discriminative Region Scoring for Fine-Grained Classification
}

\author{%
  Yujie Zhu \\
  State University of New York at Buffalo \\
  \texttt{yzhu68\string@buffalo.edu}
}

\begin{document}

\maketitle

\begin{abstract}Recent vision models such as CLIP and SAM enable training-free segmentation and semantic encoding for fine-grained classification.  A common approach is to compare the representations of segmented image regions with the text prompt embeddings of the corresponding labels. However, it remains unclear how different local regions and CLIP-based scoring
strategies affect the selection of discriminative evidence, especially when ground-truth labels are unavailable. In this paper, we propose a unified CLIP-guided label-free region scoring framework for fine-grained classification. The framework evaluates cosine similarity-based, margin-based, and entropy-based scoring strategies using both SAM-generated masks and random crops, and introduces two label-free pseudo-label variants based on global image embeddings and local region embeddings. We conduct experiments on five fine-grained classification datasets to systematically compare different region generation methods and scoring strategies. The results show that Soft Negative Margin scoring achieves the strongest performance, and pseudo-label scoring closely approximates true-label performance. Although SAM produces semantically meaningful masks, random-crop-based pseudo-label scoring consistently outperforms SAM-based scoring across all datasets, suggesting that random crops preserve surrounding information and provide more stable semantic context when pseudo-labels are noisy. In addition, SAM masks benefit from aggregating embeddings from all regions, whereas random crops tend to perform better with a smaller top-$k$ subset. These findings provide new insights for fine-grained classification.
\end{abstract}

\section{Introduction}

Fine-grained classification is a highly challenging task in image classification\cite{He_2017_CVPR}. In traditional classification problems, different categories exhibit distinct features in images, making it easier for models to learn discriminative features. Fine-grained classification, however, focuses on both overall structure and subtle differences within specific regions; these categories typically consist of hundreds of sub-classes nested under a single broad category, such as different bird species or distinct car models\cite{Yang_2018_ECCV}. Due to the high visual similarity among these images, domain experts are often required to manually distinguish and annotate between the different categories. Although the field of computer vision has made tremendous progress in recent years, thanks to advancements in deep learning\cite{10.1145/3065386}\cite{long2015fullyconvolutionalnetworkssemantic}, performance in the realm of fine-grained classification remains less than satisfactory. This challenge is largely attributable to the difficulty in localizing information-rich regions within an image and extracting discriminative features from them.
Traditional approaches to fine-grained classification have focused on localizing specific discriminative points within an image; some studies employ unsupervised methods to identify these information sources\cite{zhang2016picking}\cite{zhao2017diversified}\cite{zheng2017learning}, while others rely on learning bounding boxes or keypoints\cite{xie2013hierarchical}\cite{zhang2014part}. Although these methods can achieve impressive performance, they either rely heavily on manually annotated training data or possess complex architectures\cite{chou2023finegrainedvisualclassificationhightemperature}.
\par
In recent years, driven by the rapid development of large-scale vision-language and segmentation foundation models, large pre-trained vision-language models such as CLIP have demonstrated exceptional capabilities in zero-shot learning tasks through massive-scale training \cite{radford2021learningtransferablevisualmodels}. However, since CLIP mainly relies on global image features, it may fail to capture the subtle localized cues necessary to distinguish between fine-grained categories \cite{Parcalabescu_2022}\cite{yuksekgonul2023visionlanguagemodelsbehavelike}. Consequently, feature vectors for sub-classes within the same broad category may be highly similar, thereby increasing the difficulty of fine-grained classification. To extract semantic information from localized visual elements, some studies have explored methods from random image crops \cite{Ali_2026_WACV}\cite{li2024visualtextcrossalignmentrefining}, while others use segmentation-based regions such as SAM-generated masks\cite{Bao2026ScrapSAMCLIPAF}\cite{Aleem_2024_CVPR}. These two types of local regions provide different forms of visual evidence: random crops preserve rectangular image context, whereas SAM masks provide more structured visual evidence\cite{kirillov2023segment}. Building upon this foundation, we propose a unified CLIP-guided local region scoring framework for fine-grained classification. We first generate local candidate regions using either SAM masks or random crops, and employ CLIP to extract embeddings for local regions, the entire image, and class text. Then, we evaluate several scoring strategies to
estimate the discriminative information contained in each local region and select the top-$k$ most discriminative regions. The aggregated features are fed into a lightweight linear classifier, enabling a controlled comparison between crop-based and mask-based regions using only frozen CLIP representations, without any task-specific training or complex localization modules

\begin{figure}[h]
  \centering
   \includegraphics[width=0.9\linewidth]{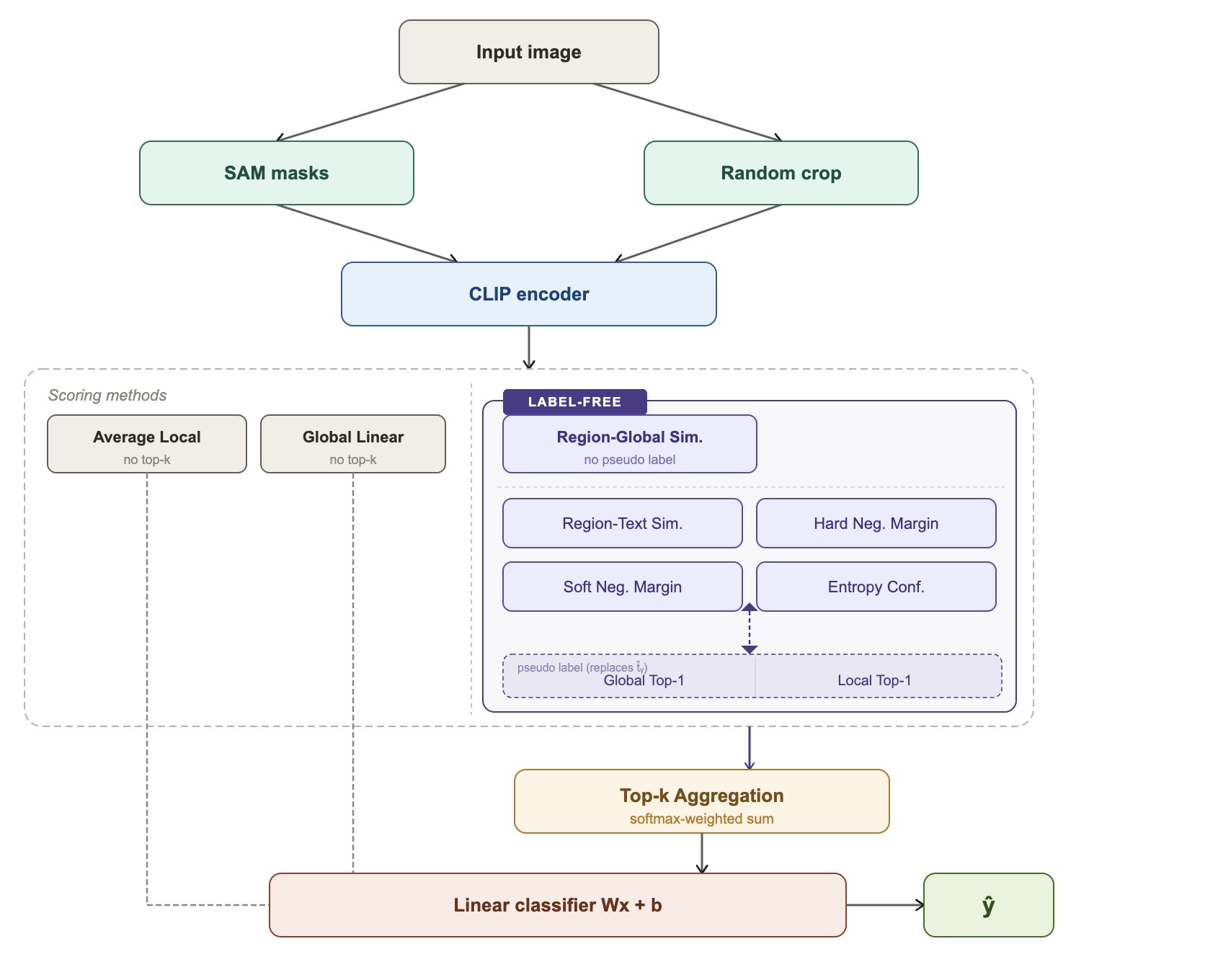}

   \caption{Overview of the proposed CLIP-guided local region scoring framework
   }
   \label{fig:pretrain}
\end{figure}


This work contributes in three primary ways:
(1) We propose multiple discriminative scoring strategies beyond cosine similarity, including hard negative margin, soft negative margin, and entropy confidence, which enhance the ability to discriminate visually similar categories.
(2) We introduce a label-free variant using pseudo labels derived from CLIP's zero-shot predictions, enabling discriminative region scoring without ground-truth supervision
(3) Through extensive experiments on multiple fine-grained classification datasets, we show that random crop regions consistently outperform SAM-based regions, with soft negative margin scoring achieving the strongest discriminative signal and global pseudo labels proving more reliable than per-region local predictions

\section{Related work}
In this section, I will briefly describe some recent related works in the areas of \textit{CLIP, SAM} and \textit{fine-grained classification}  

Li et al. (2024) \cite{li2024clipsamclipsamcollaboration}proposed a method that combines CLIP and SAM to detect subtle anomalies in images. Li et al. use CLIP to detect anomalous regions first and then apply SAM to segment them into fine-grained crops, thereby achieving the goal of localizing small targets. This work demonstrates the complementarity of CLIP and SAM and their great potential in image classification.

Ali et al. (2026) \cite{Ali_2026_WACV} proposed a method that uses CLIP to extract local crops and performs semantic alignment over these crops with CLIP to assist fine-grained classification. Ali et al.'s work shows that local cues are important for fine-grained recognition, but they do not introduce a structured way of segmenting the image, and only use cosine similarity for comparison.

Zhong et al. (2025) \cite{11356599} proposed a SAM-CLIP model for medical image classification. They use SAM to segment the image and then use CLIP to align the masks with anatomical text. However, this method does not explicitly explain how to select and which masks are more important. 

Inspired by these works, we propose a CLIP-based model for fine-grained classification and explicitly introduce several scoring methods for ranking the segmented masks, in order to assist fine-grained classification tasks.

\section{Methodology}
\subsection{Preliminaries}
The goal of the proposed CLIP-guided discriminative region scoring framework is to systematically evaluate how different local candidate generation strategies and scoring strategies affect the selection of discriminative localized visual evidence.
The overall pipeline is shown as follows:

\[
x \rightarrow \{l_i\}_{i=1}^{L}
\rightarrow \big(g,\{f_i\}_{i=1}^{L},\{t_c\}_{c=1}^{C}\big)
\rightarrow \{s_i\}_{i=1}^{L}
\xrightarrow{\mathrm{Top}(k)}
r
\rightarrow \hat{y}.
\]
where \(x\) denotes the input image, 
\(\{l_i\}_{i=1}^{L}\) denotes the set of generated local regions from either SAM-generated masks or random crops,  \(g\) denotes the CLIP embedding of the entire image, \(f_i\) denotes the local CLIP image embedding, \(t_c\) denotes the CLIP text embedding of class \(c\),  \(s_i\) is the score from scoring methods, \(r\) is the aggregated local representation and \(\hat{y}\) is the final prediction.

\subsection{Local Region Generation}
In this work, a local region is generated either from a SAM-generated mask or from a random rectangular crop. To make the two types of methods comparable, both use the same maximum number of regions and the same area range. Regions with a low area ratio typically capture noise rather than semantic content, while large regions of the image area approach global context and provide little localized discriminative evidence.
\subsubsection{SAM Mask}
Given an input image $x$, we apply SAM to generate a set of candidate masks:
\[
\{l_1, l_2, \dots, l_L\},
\]
where $L$ is the number of masks produced for that image.
\subsubsection{Random Crop}
Given an input image x, we randomly sample a set of rectangular crop regions:
\[
\{l_1, l_2, \dots, l_L\},
\]
\subsection{CLIP Feature Extraction}
Now, we use a frozen CLIP model to extract global image features, local candidate features, and class text embeddings. For the input image $x$, we compute a global image feature:
\[
g=\phi_{\text{img}}(x)\in \mathbb{R}^d,
\]
where $\phi_{\text{img}}(\cdot)$ denotes the CLIP image encoder.

Similarly, for each candidate region $l_i$, we compute a region-level feature:
\[
f_i=\phi_{\text{img}}(l_i)\in \mathbb{R}^d,\quad i=1,\dots,L.
\]

For each category $c$, we construct a text prompt of the form ``a photo of a \{class name\},'' and use the CLIP text encoder $\phi_{\text{text}}(\cdot)$ to obtain the corresponding text feature:
\[
t_c=\phi_{\text{text}}(y_c)\in \mathbb{R}^d,\quad c=1,\dots,C.
\]

All image and text features are $\ell_2$-normalized before similarity computation:
\[
\hat{f}_i=\frac{f_i}{\|f_i\|_2}, \qquad
\hat{g}=\frac{g}{\|g\|_2}, \qquad
\hat{t}_c=\frac{t_c}{\|t_c\|_2}.
\]

With $\ell_2$-normalized features, cosine similarity reduces to a dot product. These normalized features provide a common semantic embedding space in which localized visual information, global image semantics, and labeled category descriptions can be directly compared.

\subsection{Region Scoring Strategies}

\paragraph{I. Global Linear Probe}
As the most direct baseline, the global CLIP image feature of the entire image is fed directly into the linear classifier without any region-level localization:
\begin{equation}
    r = \hat{g}
\end{equation}
This baseline evaluates the contribution of global CLIP representations alone, without any local region selection.

\paragraph{II. All-Region Average.}
As a sanity-check baseline, all CLIP-encoded region features are averaged without semantic scoring, producing an aggregated local representation \(r\):
\begin{equation}
r\ = \frac{1}{M}\sum_{i=1}^{M} \hat{f}_i.
\end{equation}

\paragraph{III. Region-Global Similarity}
The region is scored by its cosine similarity to the global image embedding:
\begin{equation}
    s_i^{\text{global}} = \cos(\hat{f}_i, \hat{g})
\end{equation}

\paragraph{Label-Free Pseudo Label Strategies.}
The following four scoring strategies estimate region discriminability using a target class 
embedding $\hat{t}_y$.  We replace the ground-truth label embedding $\hat{t}_y$ with a pseudo-label embedding $\hat{t}_{\tilde{y}}$:
\par
\textit{Global pseudo label}
\par The pseudo class is the top-1 prediction based on the 
global image embedding $\hat{g}$:
\begin{equation}
    \tilde{y} = \arg\max_{c \in \{1,\dots,C\}} \cos(\hat{g}, \hat{t}_c),
\end{equation}
\par \textit{Local pseudo label}
\par Each region independently predicts its own pseudo class:
\begin{equation}
    \tilde{y}_i = \arg\max_{c \in \{1,\dots,C\}} \cos(\hat{f}_i, \hat{t}_c).
\end{equation}
In both cases, the scoring strategies require no ground-truth supervision.

\paragraph{IV. Region-Text Similarity}
The simplest semantic score measures how well a region aligns with the target class text embedding:
\[
s_i^{\text{local}} = \cos(\hat{f}_i, \hat{t}_{\tilde{y}}).
\]

This score directly evaluates whether the region is semantically similar to the target category. It serves as one of the most widely adopted scoring criteria in CLIP-based fine-grained classification.

\paragraph{V. Hard Negative Margin}
In fine-grained classification, a useful region should not only match the correct class, but also help distinguish it from visually similar competing classes. Therefore, we define a local margin score as follows:
\[
s_i^{\text{margin}}
=
\cos(\hat{f}_i, \hat{t}_{\tilde{y}})
-
\max_{c \neq y} \cos(\hat{f}_i, \hat{t}_c).
\]

This score emphasizes inter-class discriminability by penalizing regions that also align strongly with other classes.

\paragraph{VI. Soft Negative Margin}
The hard negative margin may be sensitive to a single competing class. To obtain a smoother and more robust discriminative score, we replace the single hardest negative with a weighted aggregation over all non-target classes. For each region \(i\), the negative-class weights are computed as follows:
\[
\alpha_{i,c}
=
\frac{
\exp(\tau \cos(\hat{f}_i,\hat{t}_c))
}{
\sum_{c' \ne \tilde{y}}
\exp(\tau \cos(\hat{f}_i,\hat{t}_{c'}))
},
\qquad c \ne \tilde{y},
\]
where \(\tau\) is a temperature parameter controlling the sharpness of the negative-class weighting. The soft negative margin score is then defined as follows:
\[
s_i^{\mathrm{soft}}
=
\cos(\hat{f}_i,\hat{t}_{\tilde{y}})
-
\sum_{c \ne \tilde{y}}
\alpha_{i,c}
\cos(\hat{f}_i,\hat{t}_c).
\]
This score encourages regions that are strongly aligned with the target class while maintaining separation from a weighted combination of competing classes, making the score more robust to noisy or ambiguous negative classes.

\paragraph{VII. Entropy Confidence}
The class distribution induced by each region is first computed using CLIP image-text similarities:
\[
p_i(c)
=
\frac{
\exp(\tau \cos(\hat{f}_i,\hat{t}_c))
}{
\sum_{c'=1}^{C}
\exp(\tau \cos(\hat{f}_i,\hat{t}_{c'}))
}.
\]
The entropy of this distribution is computed as follows:
\[
H(p_i)
=
-
\sum_{c=1}^{C}
p_i(c)\log p_i(c).
\]
The entropy confidence score is then defined as follows:
\[
s_i^{\mathrm{ent}}
=
p_i(\tilde{y})
-
\lambda H(p_i),
\]
where \(\lambda\) controls the strength of the entropy penalty. This score favors regions where the model is confident, filtering out ambiguous areas.

\subsection{Top-$k$ Region Aggregation}
After computing the score $s_i$ for each candidate region, we sort all candidate regions according to their scores and retain the top-$k$ regions. We denote $\mathcal{S}$ as the set of these selected regions. For each selected region $i \in \mathcal{S}$, let $f_i \in \mathbb{R}^d$ be its CLIP image feature. 

The selected regions are then aggregated using softmax-normalized importance weights $w_i$ for region $i$:
\[
w_i=\frac{\exp({s}_i)}{\sum_{j\in\mathcal{S}}\exp({s}_j)}, \qquad i\in\mathcal{S},
\]
\[
r=\sum_{i\in\mathcal{S}} w_i f_i,
\]
 This formulation ensures that higher-scoring regions contribute more strongly to the final local representation.
\subsection{Classification Head and Training Objective}
The aggregated local representation
$r$ is concatenated with the global image embedding $\hat{g}$ to form the final feature:
\begin{equation}
    \hat{r} = [\hat{g}; r] \in \mathbb{R}^{2d},
\end{equation}
where \([\cdot;\cdot]\) denotes feature concatenation.

Now, the feature $\hat{r}$ is fed into a lightweight linear classifier:
\[
z = W\hat{r} + b,
\]
where $W$ and $b$ are learnable classifier parameters, and $z \in \mathbb{R}^{C}$ is the output logit vector over all classes.

The classifier is trained with the standard cross-entropy loss:
\[
\mathcal{L}_{\text{cls}}
=
-\log
\frac{\exp(z_y)}
{\sum_{c=1}^{C}\exp(z_c)}.
\]

It can be seen that our method does not train complex localization modules or part detectors. Instead, it relies on frozen CLIP representations and simple region generation strategies. Only the linear classifier parameters are learned, enabling a controlled evaluation of each scoring strategy's contribution.

\section{Experiments}

\subsection{Datasets}

\textbf{Datasets:} We evaluate the proposed framework on five widely used fine-grained classification benchmarks: CUB-200-2011~\cite{cub}, Oxford 102 Flowers~\cite{Nilsback08}, Oxford-IIIT Pets~\cite{parkhi12a}, Stanford Cars~\cite{Krause_2013_ICCV_Workshops}, and FGVC Aircraft \cite{maji2013finegrainedvisualclassificationaircraft}. 

CUB-200-2011 focuses on fine-grained bird species recognition with 200 categories and 11,788 images, including part and attribute annotations. Oxford 102 Flowers contains 102 flower categories with large variations in scale, pose, and lighting. Oxford-IIIT Pets includes 37 cat and dog breed categories with roughly 200 images per class and substantial appearance variation. Stanford Cars consists of 16,185 images over 196 fine-grained vehicle classes, typically defined by make, model, and year. FGVC-Aircraft contains 10,000 images from 100 aircraft model variants, where categories often differ by subtle structural details.

\textbf{Implementation Details:}
We adopt CLIP ViT-B/32 for visual and textual feature extraction and use SAM ViT-B to generate candidate local regions. All CLIP embeddings are L2-normalized before computing image-text similarities. The resulting local representation is concatenated with the global CLIP image embedding and passed to a lightweight linear classifier. The classifier is trained for 30 epochs with a batch size of 32 using AdamW, with a learning rate of \(1\times10^{-3}\), weight decay of \(1\times10^{-4}\), and dropout of 0.1.

\subsection{Scoring Strategies comparison}

Table 1 compares the accuracy of five datasets under different image segmentation methods, different scoring methods, and two different pseudo-labels. In general, the Soft Negative Margin Strategy is strongest among label-free scores. It suggests that effective discriminative regions should not only align with the target class but also maintain separation from visually similar categories. Furthermore, counterintuitively, although SAM can accurately segment semantically meaningful regions, random crop performs better than SAM mask on all datasets. This may be because SAM masks lack surrounding context, resulting in poor clip representation performance on the mask. Also, SAM can produce masks that isolate pure background noise. Once selected, this significantly degrades overall model performance.


\begin{table*}[t]
\centering
\scriptsize
\setlength{\tabcolsep}{4pt}
\renewcommand{\arraystretch}{1.15}
\caption{Comparison of global baselines, average local baselines, and label-free region scoring methods on five fine-grained classification datasets with $k$ = 5.}
\label{tab:main_results}
\resizebox{\textwidth}{!}{
\begin{tabular}{lllccccc}
\toprule
Candidate & Pseudo Label & Scoring Strategy
& CUB-200-2011
& Oxford 102 Flowers
& Oxford-IIIT Pets
& Stanford Cars
& FGVC-Aircraft \\
\midrule

-- & -- & Global Linear
& 0.4558 & 0.8827 & 0.8633 & 0.7746 & 0.4551 \\

\midrule

\multirow{10}{*}{SAM Mask}
& \multirow{2}{*}{None}
& Average Local
& 0.4461 & 0.8741 & 0.8514 & 0.7683 & 0.4455 \\
& 
& Region-Global Sim.
& 0.4575 & 0.8912 & 0.8560 & 0.7513 & 0.4368 \\

\cmidrule(lr){2-8}

& \multirow{4}{*}{Global}
& Region-Text Sim.
& 0.4553 & 0.8644 & 0.8636 & 0.7586 & 0.4380 \\
&
& Hard Neg. Margin
& 0.4435 & 0.8567 & 0.8595 & 0.7553 & 0.4320 \\
&
& Soft Neg. Margin
& 0.4543 & 0.8627 & 0.8707 & 0.7568 & 0.4347 \\
&
& Entropy Confidence
& 0.4517 & 0.8600 & 0.8690 & 0.7543 & 0.4386 \\

\cmidrule(lr){2-8}

& \multirow{4}{*}{Local}
& Region-Text Sim.
& 0.4560 & 0.8592 & 0.8514 & 0.7534 & 0.4395 \\
&
& Hard Neg. Margin
& 0.4457 & 0.8520 & 0.8480 & 0.7549 & 0.4389 \\
&
& Soft Neg. Margin
& 0.4519 & 0.8475 & 0.8490 & 0.7608 & 0.4410 \\
&
& Entropy Confidence
& 0.4564 & 0.8466 & 0.8476 & 0.7533 & 0.4404 \\

\midrule

\multirow{10}{*}{Random Crop}
& \multirow{2}{*}{None}
& Average Local
& 0.6250 & 0.8987 & 0.8694 & 0.8022 & 0.4851 \\
&
& Region-Global Sim.
& 0.6282 & 0.9093 & 0.8861 & 0.8179 & 0.4956 \\

\cmidrule(lr){2-8}

& \multirow{4}{*}{Global}
& Region-Text Sim.
& 0.6535 & 0.9133 & 0.8949 & 0.8165 & 0.4887 \\
&
& Hard Neg. Margin
& 0.6522 & 0.9071 & 0.8901 & 0.8047 & 0.4815 \\
&
& Soft Neg. Margin
& \best{0.6559} & 0.9138 & \best{0.8976} & 0.8183 & 0.5017 \\
&
& Entropy Confidence
& 0.6115 & \best{0.9149} & 0.8946 & 0.8268 & 0.5083 \\

\cmidrule(lr){2-8}

& \multirow{4}{*}{Local}
& Region-Text Sim.
& 0.6085 & 0.9076 & 0.8850 & 0.8177 & 0.4983 \\
&
& Hard Neg. Margin
& 0.5910 & 0.9014 & 0.8827 & 0.8088 & 0.4836 \\
&
& Soft Neg. Margin
& 0.6040 & 0.9096 & 0.8912 & \best{0.8271} & \best{0.5164} \\
&
& Entropy Confidence
& 0.6038 & 0.9096 & 0.8898 & 0.8182 & 0.5155 \\

\bottomrule
\end{tabular}
}
\end{table*}
\par
Table 2 compares pseudo-label-based scoring with true-label scoring. It is shown that pseudo-label scoring closely matches true-label performance, indicating that pseudo labels derived from global predictions are sufficiently reliable for guiding region selection.

\begin{table*}[t]
\centering
\scriptsize
\setlength{\tabcolsep}{3.2pt}
\renewcommand{\arraystretch}{1.12}
\caption{Comparison between true-label scoring and pseudo-label scoring on five fine-grained classification datasets. For each dataset, \textbf{True} denotes the test accuracy of the true-label method, $\Delta_G=\text{Pseudo(Global)}-\text{True}$, and $\Delta_L=\text{Pseudo(Local)}-\text{True}$. }
\label{tab:true_vs_pseudo}
\resizebox{\textwidth}{!}{
\begin{tabular}{llccc ccc ccc ccc ccc}
\toprule
\multirow{2}{*}{Candidate} & \multirow{2}{*}{Scoring Strategy}
& \multicolumn{3}{c}{CUB-200-2011}
& \multicolumn{3}{c}{Oxford 102 Flowers}
& \multicolumn{3}{c}{Oxford-IIIT Pets}
& \multicolumn{3}{c}{Stanford Cars}
& \multicolumn{3}{c}{FGVC-Aircraft} \\
\cmidrule(lr){3-5}\cmidrule(lr){6-8}\cmidrule(lr){9-11}\cmidrule(lr){12-14}\cmidrule(lr){15-17}
&
& True & $\Delta_G$ & $\Delta_L$
& True & $\Delta_G$ & $\Delta_L$
& True & $\Delta_G$ & $\Delta_L$
& True & $\Delta_G$ & $\Delta_L$
& True & $\Delta_G$ & $\Delta_L$ \\
\midrule

\multirow{4}{*}{SAM Mask}
& Region-Text Sim.
& 0.5873 & -0.1320 & -0.1313
& 0.8675 & \best{-0.0031} & -0.0083
& 0.8731 & \best{-0.0095} & -0.0217
& 0.7651 & \best{-0.0065} & -0.0117
& 0.4380 & \best{+0.0000} & +0.0015 \\

& Hard Neg. Margin
& 0.6516 & -0.2081 & -0.2059
& 0.8850 & -0.0283 & -0.0330
& 0.8925 & -0.0330 & -0.0445
& 0.8005 & -0.0452 & -0.0456
& 0.4986 & -0.0666 & -0.0597 \\

& Soft Neg. Margin
& 0.7002 & -0.2459 & \worst{-0.2483}
& 0.8966 & -0.0339 & \worst{-0.0491}
& 0.8939 & -0.0232 & -0.0449
& 0.7988 & -0.0420 & -0.0380
& 0.5062 & -0.0715 & -0.0652 \\

& Entropy Confidence
& 0.5747 & -0.1230 & -0.1183
& 0.8826 & -0.0226 & -0.0360
& 0.8963 & -0.0273 & \worst{-0.0487}
& 0.7723 & -0.0180 & -0.0190
& 0.4926 & -0.0540 & -0.0522 \\

\midrule

\multirow{4}{*}{Random Crop}
& Region-Text Sim.
& 0.6942 & -0.0407 & -0.0857
& 0.9218 & -0.0085 & -0.0142
& 0.9068 & -0.0119 & -0.0218
& 0.8470 & -0.0305 & -0.0293
& 0.5392 & -0.0505 & -0.0409 \\

& Hard Neg. Margin
& 0.6705 & -0.0183 & -0.0795
& 0.9294 & -0.0223 & -0.0280
& 0.9218 & -0.0317 & -0.0391
& 0.8671 & \worst{-0.0624} & -0.0583
& 0.5896 & \worst{-0.1081} & -0.1060 \\

& Soft Neg. Margin
& 0.6872 & -0.0313 & -0.0832
& 0.9228 & -0.0090 & -0.0132
& 0.9099 & -0.0123 & -0.0187
& 0.8512 & -0.0329 & -0.0241
& 0.5566 & -0.0549 & -0.0402 \\

& Entropy Confidence
& 0.6233 & \best{-0.0118} & -0.0195
& 0.9193 & -0.0044 & -0.0097
& 0.9054 & -0.0108 & -0.0156
& 0.8346 & -0.0078 & -0.0164
& 0.5341 & -0.0258 & -0.0186 \\

\bottomrule
\end{tabular}
}
\end{table*}


\begin{figure*}[t]
    \centering
    \includegraphics[width=\textwidth]{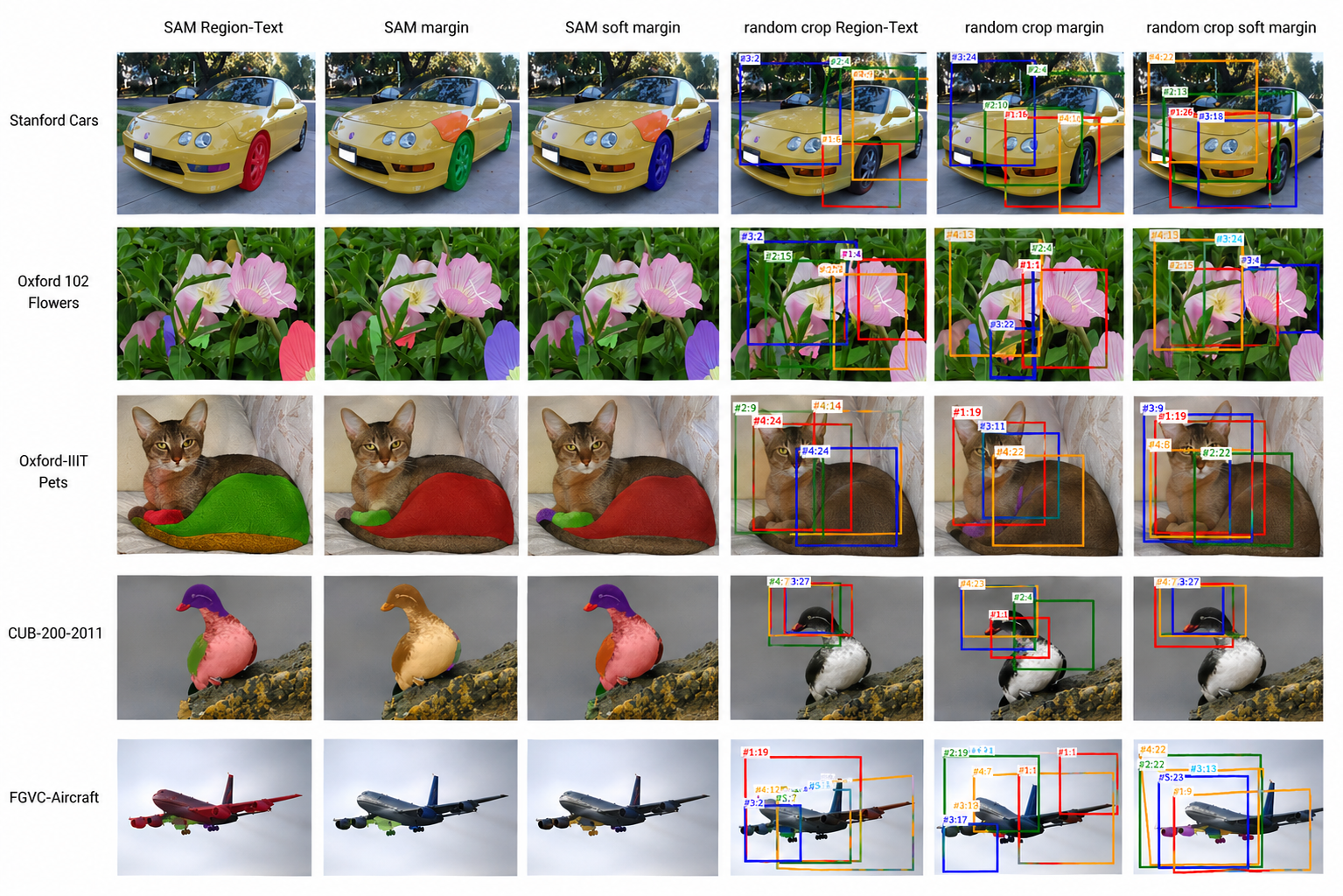}
    \caption{
    Qualitative visualization of selected local regions across five fine-grained classification datasets.
    }
    \label{fig:qualitative_region_visualization}
\end{figure*}

\subsection{Top k}

Table 3 analyzes the impact of varying choices of $k$ on classification performance. We observed that the model's performance remained relatively stable across a wide range of $k$ values, indicating that the method is not highly sensitive to the specific choice of $k$. However, interestingly, when using all regions, the accuracy of random cropping significantly decreased, while the accuracy of SAM masking increased. We believe this is because lower-ranked random crops contain more noise, thus affecting overall performance, while full SAM masks contain more contextual information, thereby improving accuracy.

\begin{table*}[t]
\centering
\scriptsize
\setlength{\tabcolsep}{4pt}
\renewcommand{\arraystretch}{1.12}
\caption{Top-$k$ sensitivity analysis for random-crop-based pseudo-label scoring methods.}
\label{tab:crop_topk_sensitivity}
\resizebox{\textwidth}{!}{
\begin{tabular}{llccccc}
\toprule
Scoring Strategy & Top-$k$
& CUB-200-2011
& Oxford 102 Flowers
& Oxford-IIIT Pets
& Stanford Cars
& FGVC-Aircraft \\
\midrule

\multirow{7}{*}{Soft Neg. Margin (Global)}
& 2 & 0.6492 & \textbf{0.9174} & 0.8976 & 0.8205 & 0.4992 \\
& 3 & 0.6469 & 0.9130 & 0.8939 & 0.8203 & 0.5032 \\
& 4 & 0.6574 & 0.9141 & 0.8949 & 0.8223 & 0.5011 \\
& 5 & 0.6488 & 0.9151 & \textbf{0.8980} & \textbf{0.8243} & \textbf{0.5050} \\
& 6 & \textbf{0.6580} & 0.9125 & 0.8946 & 0.8222 & 0.5029 \\
& 7 & 0.6485 & 0.9156 & 0.8929 & 0.8214 & 0.5005 \\
& ALL & 0.6252 & 0.9014 & 0.8660 & 0.8014 & 0.4881 \\

\midrule

\multirow{7}{*}{Entropy Confidence (Global)}
& 2 & 0.6105 & 0.9156 & \textbf{0.8983} & 0.8192 & 0.5038 \\
& 3 & 0.6092 & 0.9128 & 0.8946 & 0.8271 & 0.5086 \\
& 4 & 0.6087 & 0.9153 & 0.8973 & 0.8222 & 0.5101 \\
& 5 & 0.6070 & \textbf{0.9175} & 0.8946 & \textbf{0.8284} & \textbf{0.5107} \\
& 6 & 0.6096 & 0.9078 & 0.8935 & 0.8260 & 0.5038 \\
& 7 & 0.6085 & 0.9140 & 0.8946 & 0.8269 & 0.5086 \\
& ALL & \textbf{0.6237} & 0.9001 & 0.8646 & 0.7987 & 0.4827 \\

\midrule

\multirow{7}{*}{Soft Neg. Margin (Local)}
& 2 & 0.5985 & 0.9068 & \textbf{0.9003} & 0.8260 & 0.5095 \\
& 3 & 0.6077 & 0.9099 & 0.8973 & \textbf{0.8283} & 0.5122 \\
& 4 & 0.6096 & 0.9104 & 0.8922 & 0.8260 & 0.5149 \\
& 5 & 0.6094 & \textbf{0.9120} & 0.8956 & 0.8251 & \textbf{0.5248} \\
& 6 & 0.6027 & 0.9101 & 0.8939 & 0.8281 & 0.5185 \\
& 7 & 0.6094 & 0.9094 & 0.8922 & 0.8265 & 0.5158 \\
& ALL & \textbf{0.6214} & 0.8993 & 0.8646 & 0.8007 & 0.4836 \\

\midrule

\multirow{7}{*}{Entropy Confidence (Local)}
& 2 & 0.6066 & 0.9083 & \textbf{0.8956} & 0.8219 & 0.5119 \\
& 3 & 0.6055 & 0.9097 & 0.8949 & 0.8272 & 0.5158 \\
& 4 & 0.6081 & 0.9117 & 0.8946 & \textbf{0.8292} & 0.5179 \\
& 5 & 0.6060 & \textbf{0.9119} & 0.8946 & 0.8254 & 0.5182 \\
& 6 & 0.6060 & 0.9102 & 0.8935 & 0.8269 & 0.5176 \\
& 7 & 0.6060 & 0.9099 & 0.8912 & 0.8252 & \textbf{0.5224} \\
& ALL & \textbf{0.6248} & 0.9003 & 0.8646 & 0.8010 & 0.4851 \\

\bottomrule
\end{tabular}
}
\end{table*}

\begin{table*}[t]
\centering
\scriptsize
\setlength{\tabcolsep}{4pt}
\renewcommand{\arraystretch}{1.12}
\caption{Top-$k$ sensitivity analysis for SAM-based pseudo-label scoring methods.}
\label{tab:sam_topk_sensitivity}
\resizebox{\textwidth}{!}{
\begin{tabular}{llccccc}
\toprule
Scoring Strategy & Top-$k$
& CUB-200-2011
& Oxford 102 Flowers
& Oxford-IIIT Pets
& Stanford Cars
& FGVC-Aircraft \\
\midrule

\multirow{7}{*}{Soft Neg. Margin (Global)}
& 2 & \textbf{0.4654} & 0.8463 & 0.8680 & 0.7453 & 0.4347 \\
& 3 & 0.4615 & 0.8544 & 0.8680 & 0.7526 & 0.4368 \\
& 4 & 0.4566 & 0.8610 & 0.8663 & 0.7539 & 0.4338 \\
& 5 & 0.4510 & 0.8671 & 0.8687 & 0.7566 & 0.4329 \\
& 6 & 0.4474 & 0.8678 & 0.8667 & 0.7582 & 0.4359 \\
& 7 & 0.4534 & 0.8684 & \textbf{0.8701} & 0.7569 & 0.4368 \\
& ALL & 0.4410 & \textbf{0.8738} & 0.8463 & \textbf{0.7640} & \textbf{0.4404} \\

\midrule

\multirow{7}{*}{Entropy Confidence (Global)}
& 2 & \textbf{0.4630} & 0.8434 & 0.8680 & 0.7442 & 0.4341 \\
& 3 & 0.4615 & 0.8549 & 0.8711 & 0.7477 & 0.4332 \\
& 4 & 0.4588 & 0.8595 & \textbf{0.8714} & 0.7520 & 0.4410 \\
& 5 & 0.4523 & 0.8653 & 0.8650 & 0.7476 & 0.4362 \\
& 6 & 0.4555 & 0.8696 & 0.8670 & 0.7530 & 0.4368 \\
& 7 & 0.4540 & 0.8679 & 0.8677 & 0.7540 & 0.4365 \\
& ALL & 0.4478 & \textbf{0.8754} & 0.8435 & \textbf{0.7649} & \textbf{0.4452} \\

\midrule

\multirow{7}{*}{Soft Neg. Margin (Local)}
& 2 & 0.4446 & 0.8195 & 0.8510 & 0.7500 & 0.4251 \\
& 3 & 0.4420 & 0.8380 & \textbf{0.8554} & 0.7502 & 0.4308 \\
& 4 & 0.4493 & 0.8434 & 0.8531 & 0.7576 & 0.4347 \\
& 5 & 0.4553 & 0.8492 & 0.8432 & 0.7629 & 0.4377 \\
& 6 & \textbf{0.4555} & 0.8544 & 0.8503 & 0.7597 & 0.4395 \\
& 7 & 0.4538 & 0.8567 & 0.8503 & 0.7626 & 0.4386 \\
& ALL & 0.4388 & \textbf{0.8714} & 0.8466 & \textbf{0.7648} & \textbf{0.4446} \\

\midrule

\multirow{7}{*}{Entropy Confidence (Local)}
& 2 & 0.4551 & 0.8258 & \textbf{0.8548} & 0.7493 & 0.4293 \\
& 3 & \textbf{0.4568} & 0.8333 & 0.8456 & 0.7545 & 0.4344 \\
& 4 & 0.4525 & 0.8419 & 0.8527 & 0.7559 & 0.4386 \\
& 5 & 0.4515 & 0.8520 & 0.8497 & 0.7580 & 0.4410 \\
& 6 & 0.4489 & 0.8544 & 0.8531 & 0.7619 & 0.4380 \\
& 7 & 0.4489 & 0.8556 & 0.8517 & 0.7625 & 0.4374 \\
& ALL & 0.4485 & \textbf{0.8681} & 0.8429 & \textbf{0.7657} & \textbf{0.4440} \\

\bottomrule
\end{tabular}
}
\end{table*}

\section{Limitation and Conclusion.}
Our experiments reveal that random crops outperform SAM-based regions across all five datasets, despite SAM requiring over 10 times more preprocessing time. A possible reason for this is that SAM masks lack surrounding context, limiting CLIP's ability to produce more discriminative region embeddings. Meanwhile, SAM can produce masks that isolate pure background noise. Once selected, it will significantly degrade overall model performance. Furthermore, soft negative margin scoring achieves the strongest performance, suggesting that effective discriminative regions should not only align with the target class but also maintain separation from visually similar categories.

However, our framework relies heavily on frozen CLIP  representations. As observed on FGVC-Aircraft dataset, when CLIP  representation lacks sufficient discriminability for the target domain, the overall performance degrades substantially. Future work could explore domain-adapted encoders or stronger pseudo-label strategies for challenging datasets.

\FloatBarrier
\bibliographystyle{plainnat}
\bibliography{references}

\end{document}